\documentclass{article} 
\usepackage{nips15submit_e,times}
\usepackage{hyperref}
\usepackage{url}
\usepackage[pdftex]{graphicx}
\usepackage{amsmath}
\usepackage{enumerate}
\usepackage{bibspacing}

\title{Recursive Training of 2D-3D Convolutional Networks for Neuronal Boundary Detection}

\author{
Kisuk Lee, Aleksandar Zlateski\\
Massachusetts Institute of Technology \\
\texttt{\{kisuklee,zlateski\}@mit.edu}
\And
Ashwin Vishwanathan, H. Sebastian Seung \\
Princeton University \\
\texttt{\{ashwinv,sseung\}@princeton.edu}
}

\nipsfinalcopy 

\begin{document}

\maketitle

\begin{abstract}
Efforts to automate the reconstruction of neural circuits from 3D
electron microscopic (EM) brain images are critical for the field of
connectomics. An important computation for reconstruction is the
detection of neuronal boundaries. Images acquired by serial section
EM, a leading 3D EM technique, are highly anisotropic, with inferior
quality along the third dimension. For such images, the 2D
max-pooling convolutional network has set the standard for performance
at boundary detection. Here we achieve a substantial gain in accuracy
through three innovations. Following the trend towards deeper
networks for object recognition, we use a much deeper network than
previously employed for boundary detection. Second, we incorporate 3D
as well as 2D filters, to enable computations that use 3D context.
Finally, we adopt a recursively trained architecture in which a first network
generates a preliminary boundary map that is provided as input along
with the original image to a second network that generates a final
boundary map. Backpropagation training is accelerated by ZNN, a new
implementation of 3D convolutional networks that uses multicore CPU
parallelism for speed. Our hybrid 2D-3D architecture could be more
generally applicable to other types of anisotropic 3D images,
including video, and our recursive framework for any image labeling
problem.
\end{abstract}

\section{Introduction}

Neural circuits can be reconstructed by analyzing 3D brain images from
electron microscopy (EM)~\cite{White1986}. Image analysis has been accelerated
by semiautomated systems that use computer vision to reduce the amount
of human labor required~\cite{Takemura2013,
  Helmstaedter2013,Kim2014}. However, analysis of large image datasets
is still laborious \cite{Helmstaedter2014}, so it is critical to increase
automation by improving the accuracy of computer vision algorithms.

A variety of machine learning approaches have been explored for the 3D
reconstruction of neurons, a problem that can be formulated as image
segmentation or boundary detection~\cite{Jain2010,Tasdizen2014}.  This
paper focuses on neuronal boundary detection in images from serial
section EM, the most widespread kind of 3D EM~\cite{Briggman2012}. The technique starts by cutting and collecting ultrathin ($30$ to
$100$ nm) sections of brain tissue.  A 2D image is acquired from each
section, and then the 2D images are aligned.  The spatial resolution
of the resulting 3D image stack along the $z$ direction (perpendicular
to the cutting plane) is set by the thickness of the sections. This is
generally much worse than the resolution that EM yields in the $xy$
plane.  In addition, alignment errors may corrupt the image along the
$z$ direction.

Due to these issues with the $z$ direction of the image stack
\cite{Tasdizen2014,Jurrus2010}, most existing analysis pipelines begin
with 2D processing and only later transition to 3D.  The stages are:
(1) neuronal boundary detection within each 2D image, (2)
segmentation of neuron cross sections within each 2D image, and (3) 3D
reconstruction of individual neurons by linking across multiple 2D
images~\cite{Takemura2013,Liu2014}.

\begin{figure}[!b]
\begin{center}
\includegraphics[width=0.964\textwidth]{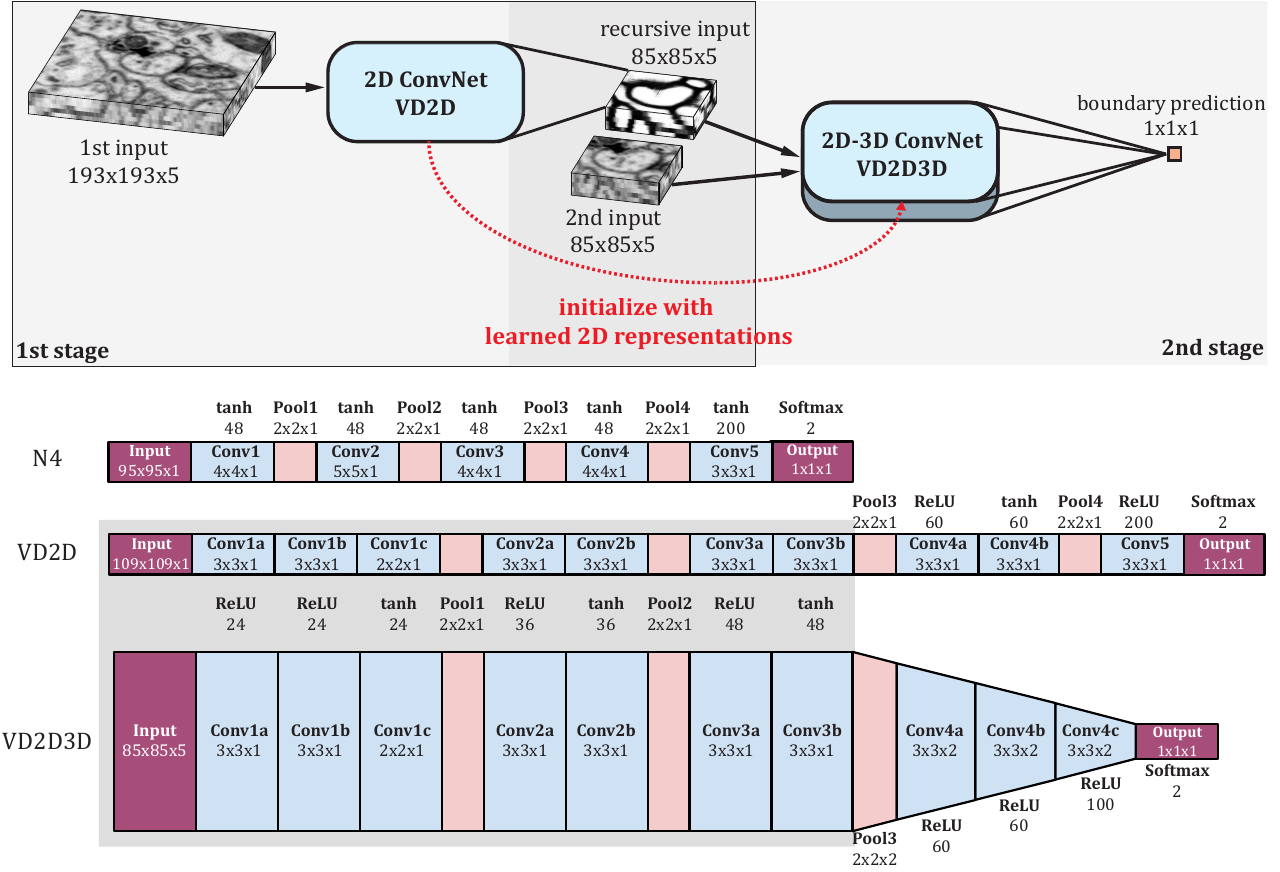}
\end{center}
\caption{An overview of our proposed framework (top) and model architectures (bottom). The number of trainable parameters in each model is $220$K (N4), $230$K (VD2D), $310$K (VD2D3D).}
\label{overview}
\end{figure}

Boundary detection in serial section EM images is done by a variety of
algorithms. Many algorithms were compared in the ISBI'12 2D EM
segmentation challenge, a publicly available dataset and benchmark~\cite{ISBI2012}.
The winning submission was an ensemble of max-pooling convolutional networks (ConvNets)
created by IDSIA \cite{Ciresan2012}. One of the
ConvNet architectures shown in Figure~\ref{overview} (N4) is the largest architecture from \cite{Ciresan2012}, and serves as a
performance baseline for the research reported here.

We improve upon N4 by adding several new elements (Fig.~\ref{overview}):
\begin{enumerate}[]
\item{\bf Increased depth}$\quad$  Our VD2D architecture is deeper than N4 (Figure~\ref{overview}), and borrows other now-standard practices from the literature, such as rectified linear units (ReLUs), small filter sizes, and multiple convolution layers between pooling layers.  VD2D already outperforms N4, without any use of 3D context. VD2D is motivated by the principle ``the deeper, the better,'' which has become popular for ConvNets applied to object recognition~\cite{Krizhevsky2012,Simonyan2015,He2015}.
\item{\bf 3D as well as 2D}$\quad$ When human experts detect boundaries in EM images, they use 3D context to disambiguate certain locations. VD2D3D is also able to use 3D context, because it contains 3D filters in its later layers. ConvNets with 3D filters were previously applied to block face EM images~\cite{Helmstaedter2013,Kim2014,Turaga2010}. Block face EM is another class of 3D EM techniques, and produces nearly isotropic images, unlike serial section EM.  VD2D3D also contains 2D filters in its earlier layers. This novel hybrid use of 2D and 3D filters is suited for the highly anisotropic nature of serial section EM images.
\item{\bf Recursive training of ConvNets}$\quad$ VD2D and VD2D3D are concatenated to create an extremely deep network. The output of VD2D is a preliminary boundary map, which is provided as input to VD2D3D in addition to the original image (Fig.~\ref{overview}). Based on these two inputs, VD2D3D is trained to compute the final boundary map.  Such ``recursive'' training has previously been applied to neural networks for boundary detection \cite{Jurrus2010,Huang2014,Seyedhosseini2013}, but not to ConvNets.
\item{\bf ZNN for 3D deep learning} Very deep ConvNets with
  3D filters are computationally expensive, so an efficient software
  implementation is critical.  We trained our networks with ZNN (\url{https://github.com/seung-lab/znn-release}), which uses multicore CPU parallelism for speed. ZNN is one of the few deep learning implementations that is well-optimized for 3D.

\end{enumerate}


While we have applied the above elements to serial section EM images,
they are likely to be generally useful for other types of images. The
hybrid use of 2D and 3D filters may be useful for video, which can
also be viewed as an anisotropic 3D image.  Previous 3D ConvNets
applied to video processing~\cite{Tran2014,Yao2015} have used 3D
filters exclusively.

Recursively trained ConvNets are potentially useful for any image labeling
problem.  The approach is very similar to recurrent ConvNets
\cite{Pinheiro2014}, which iterate the same ConvNet.  The recursive
approach uses different ConvNets for the successive iterations. The
recursive approach has been justified in several ways. In MRF/CRF
image labeling, it is viewed as the sequential refinement of the
posterior probability of a pixel being assigned a label, given both an
input image and recursive input from the previous
step~\cite{Tu2008}. Another viewpoint on recursive training is that
statistical dependencies in label (category) space can be directly
modeled from the recursive input~\cite{Huang2014}.  From the
neurobiological viewpoint, using a preliminary boundary map for an
image to guide the computation of a better boundary map for the image
can be interpreted as employing a top-down or attentional mechanism.

We expect ZNN to have applications far beyond the one considered in
this paper.  ZNN can train very large networks, because CPUs can
access more memory than GPUs. Task parallelism, rather than
the SIMD parallelism of GPUs, allows for efficient training of
ConvNets with arbitrary topology. A self-tuning capability
automatically optimizes each layer by choosing between direct and
FFT-based convolution.  FFT convolution may be more efficient for
wider layers or larger filter size
~\cite{Mathieu2014,Vasilache2015}. Finally, ZNN may incur less
software development cost, owing to the relative ease of the
general-purpose CPU programming model.

Finally, we applied our ConvNets to images from a new serial section
EM dataset from the mouse piriform cortex.  This dataset is important
to us, because we are interested in conducting neuroscience research
concerning this brain region.  Even to those with no interest in
piriform cortex, the dataset could be useful for research on image
segmentation algorithms.  Therefore we will make the annotated dataset
publicly available pending acceptance of this paper.



\begin{figure}[b]
\begin{center}
\includegraphics[width=1.0\textwidth]{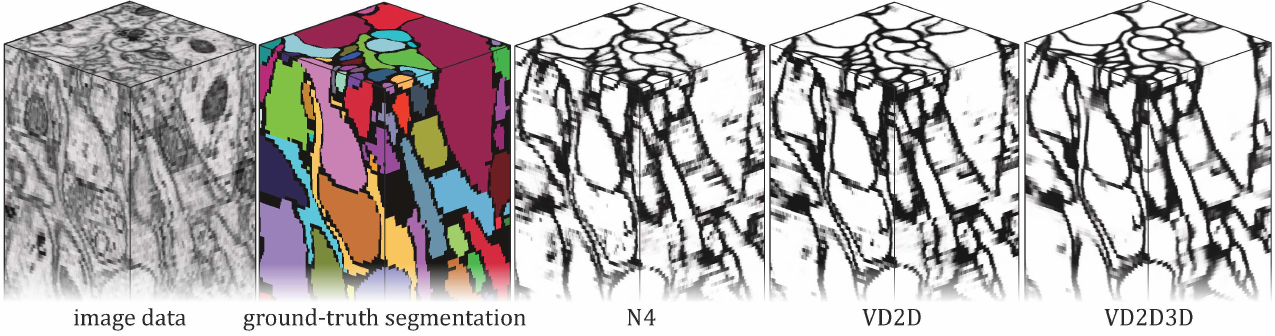}
\end{center}
\caption{Example dataset (\texttt{stack1}, Table~\ref{tab:Datasets}) and results of each architecture on \texttt{stack1}}
\label{dataset}
\end{figure}

\section{Dataset and evaluation}

{\bf Images of mouse piriform cortex}$\quad$
The datasets described here were
acquired from the piriform cortex of an adult mouse prepared with aldehyde fixation and reduced osmium staining~\cite{Tapia2012}.
The tissue was sectioned using the automatic tape collecting
ultramicrotome (ATUM)\cite{Kasthuri2015} and sections were imaged on a
Zeiss field emission scanning electron microscope~\cite{Hayworth2014}.
The 2D images were assembled into 3D stacks using custom MATLAB
routines and TrakEM2, and each stack was manually annotated using
VAST~\cite{Kasthuri2015} (Figure~\ref{dataset}). Then each stack was checked and
corrected by another annotator.

The properties of the four image stacks are detailed in Table
\ref{tab:Datasets}. It should be noted that image quality varies
across the stacks, due to aging of the field emission source in the
microscope.  In all experiments we used \texttt{stack1} for testing,
\texttt{stack2} and \texttt{stack3} for training, and \texttt{stack4}
as an additional training data for recursive training.

\begin{table}[t]
\caption{Piriform cortex datasets
\label{tab:Datasets}}
\begin{center}
\begin{tabular}{|l|cccc|}
\hline
Name    &\texttt{stack1}   &\texttt{stack2}    &\texttt{stack3}    &\texttt{stack4}
\\ \hline
Resolution (nm$^3$)     &$7\cdot7\cdot40$ &$7\cdot7\cdot40$ &$7\cdot7\cdot40$  &$10\cdot10\cdot40$
\\ 
Dimension (voxel$^3$) &$255\cdot255\cdot168$ &$512\cdot512\cdot170$ &$512\cdot512\cdot169$ &$256\cdot256\cdot121$
\\ 
\# samples & $10.9$M & $44.6$ M & $44.3$ M & $7.9$ M
\\
Usage & Test & Training & Training & Training (extra)
\\ \hline
\end{tabular}
\end{center}
\end{table}

{\bf Pixel error}$\quad$
We use softmax activation in the output layer of our nets to produce per-pixel real-valued outputs between $0$ and $1$, each of which is interpreted as the probability of an output pixel being boundary, or vice versa. This real-valued ``boundary map'' can be thresholded to generate a binary boundary map, from which the pixel-wise classification error is computed. We report the best classification error obtained by optimizing the binarization threshold with line search.

{\bf Rand score}$\quad$
We evaluate 2D segmentation performance with the Rand scoring system \cite{Rand1971,Unnikrishnan2007}. Let $n_{ij}$ denote the number of voxels simultaneously in the $i^\text{th}$ segment of the proposal segmentation and the $j^\text{th}$ segment of the ground truth segmentation. The Rand merge score and the Rand split score \begin{align*}
  V^\text{Rand}_\text{merge} = \frac{\sum_{ij}n_{ij}^2}{\sum_i(\sum_j n_{ij})^2},\quad
  V^\text{Rand}_\text{split} = \frac{\sum_{ij}n_{ij}^2}{\sum_j(\sum_i n_{ij})^2}.
\end{align*}
are closer to one when there are fewer merge and split errors, respectively. The Rand F-score is the harmonic mean of $V^\text{Rand}_\text{merge}$ and $V^\text{Rand}_\text{split}$.


To compute the Rand scores, we need to first obtain 2D neuronal segmentation based on the real-valued boundary map. To this end, we apply two segmentation algorithms with different levels of sophistication: (1) simple thresholding followed by computing 2D connected components, and (2) modified graph-based watershed algorithm~\cite{Zlateski2015}. We report the best Rand F-score obtained by optimizing parameters for each algorithm with line search, as well as the precision-recall curve for the Rand scores.


\section{Training with ZNN}
ZNN was built for 3D ConvNets. 2D convolution is regarded as a special
case of 3D convolution, in which one of the three filter dimensions
has size $1$. How ZNN implements task parallelism on multicore CPUs
will be described elsewhere. Here we describe only aspects of ZNN that are
helpful for understanding how it was used to implement the ConvNets of this paper.

{\bf Dense output with maximum filtering}$\quad$
In object recognition, a ConvNet is commonly applied to produce a
single output value for an entire input image.  However, there are
many applications in which dense output is required, i.e., the ConvNet
should produce an output image with the same resolution as the
original input image.  Such applications include boundary detection
\cite{Ciresan2012}, image labeling \cite{Long2015}, and object
localization \cite{Sermanet2014}.

\begin{figure}[t]
\begin{center}
\includegraphics[width=0.91\textwidth]{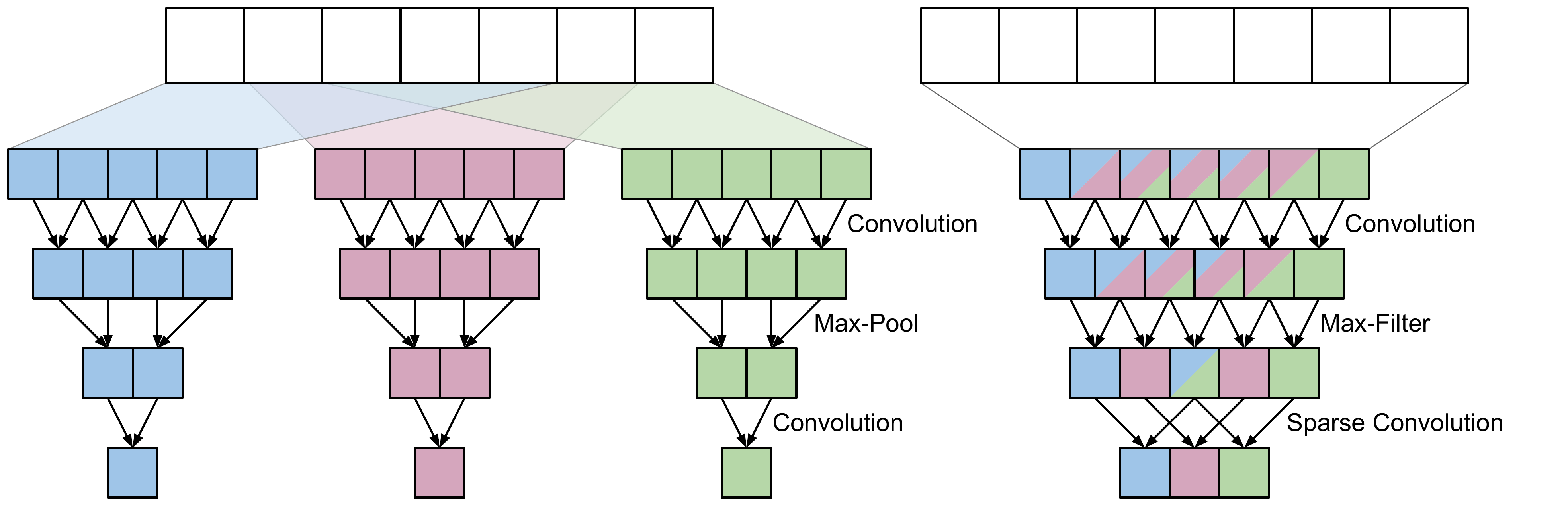}
\end{center}
\caption{Sliding window max-pooling ConvNet (left) applied on three color-coded adjacent input windows producing three outputs. Equivalent outputs produced by a max-filtering ConvNet with sparse filters (right) applied on a larger window. Computation is minimized by reusing the intermediate values for computing multiple outputs (as color coded).}
\label{max-filter}
\end{figure}

ZNN was built from the ground up for dense output and also for dense
feature maps.\footnote{Feature maps with the same resolution as the original input image.
See Figure~\ref{feature} for example. Note that the feature maps shown in Figure~\ref{feature}
keep the original resolution even after a couple of max-pooling layers.}
ZNN employs max-filtering, which slides a window across
the image and applies the maximum operation to the window (Figure~\ref{max-filter}).
Max-filtering is the dense variant of max-pooling.  Consequently all
feature maps remain intact as dense 3D volumes during both forward and
backward passes, making them straightforward for visualization and maipulation.

On the other hand, all filtering operations are sparse, in the sense
that the sliding window samples sparsely from a regularly spaced set
of voxels in the image (Figure~\ref{max-filter}).  ZNN can control the spacing/sparsity of any
filtering operation, either convolution or max-filtering.

ZNN can efficiently compute the dense output of a sliding window
max-pooling ConvNet by making filter sparsity depend on the number of
prior max-filterings.  More specifically, each max-filtering increases
the sparsity of all subsequent filterings by a factor equal to the
size of the max-pooling window.  This approach, which we employ for
the paper, is also called ``skip-kernels'' \cite{Sermanet2014} or
``filter rarefaction'' \cite{Long2015}, and is equivalent in its
results to
``max-fragmentation-pooling''~\cite{Giusti2013,Masci2013}. Note
however that ZNN is more general, as the sparseness of filters need
not depend on max-filtering, but can be controlled independently.

{\bf Output patch training}$\quad$
Training in ZNN is based on loss computed over a dense output patch of
arbitrary size. The patch can be arbitrarily large, limited only by memory.
This includes the case of a patch that spans the
entire image \cite{Long2015, Masci2013}.  Although large patch sizes
reduce the computational cost per output pixel, neighboring pixels in
the patch may provide redundant information.  In practice, we choose
an intermediate output patch size.


\section{Network architecture}

{\bf N4}$\quad$
As a baseline for performance comparisons, we adopted the largest 2D ConvNet architecture (named N4) from Cire\c{s}an et al.~\cite{Ciresan2012} (Figure~\ref{overview}).

{\bf VD2D}$\quad$
The architecture of VD2D (``Very Deep 2D'') is shown in
Figure~\ref{overview}.  All convolution filters are $3\times3\times1$, except that
\texttt{Conv1c} uses a $2\times2\times1$ filter to make the ``field of view'' or ``receptive field'' for a single output pixel have an odd-numbered size and therefore centerable around the output pixel.  Due to the use of smaller filters, the number of trainable parameters in VD2D is roughly the same as in the shallower N4.

Multiple convolution layers are between each max-pooling layer. Some
convolution layers have hyperbolic tangent (tanh) nonlinearities rather than ReLU. We
resorted to this design because preliminary experiments showed that 2D
ConvNets trained on either \texttt{stack2} or \texttt{stack3}
performed badly on one another.  This is probably because the two
stacks differed in image quality (Table~\ref{tab:Datasets}).  We added the
tanh layers thinking that their outputs might saturate near the asymptotes,
suppressing variation in feature map activations due to image quality
variations. We have not had the chance to verify whether this
speculation is correct, so the tanh nonlinearities may not be
important here.

{\bf VD2D3D}$\quad$
The architecture of VD2D3D (``Very Deep 2D-3D'') is initially identical to VD2D (Figure~\ref{overview}), except that later convolution layers switch to $3\times 3\times 2$ filters. This causes the number of trainable parameters to increase, so we compensate by trimming the size of \texttt{Conv4c} to just $100$ feature maps.  The 3D filters in the later layers should enable the network to use 3D context to detect neuronal boundaries.  The use of 2D filters in the initial layers makes the network faster to run and train.

{\bf Recursive training}$\quad$
It is possible to apply VD2D3D by itself to boundary detection, giving
the raw image as the only input. However, we use a recursive approach
in which VD2D3D receives an extra input, the output of VD2D. As we will see below, this produces a significant improvement in performance. It should be noted that instead of providing the recursive input directly to VD2D3D, we added new layers~\footnote{These layers are identical to \texttt{Conv1a}, \texttt{Conv1b}, and \texttt{Conv1c}.} dedicated to processing it. This separate, parallel processing stream for recursive input joins the main stream at \texttt{Conv1c}, allowing for more complex, highly nonlinear interaction between the low-level features and the contextual information in the recursive input.

\section{Training procedures}
Networks were trained using backpropagation with the cross-entropy
loss function.  We first trained VD2D, and then trained VD2D3D.  The
2D layers of VD2D3D were initialized using trained weights from VD2D.
This initialization meant that our recursive approach bore some
similarity to recurrent ConvNets, in which the first and second stage
networks are constrained to be identical \cite{Pinheiro2014}.
However, we did not enforce exact weight sharing, but trained the
weights of VD2D3D.

\begin{enumerate}[]
\item{\bf Output patch}$\quad$
As mentioned earlier, training with ZNN is done by dense output patch-based gradient update with per-pixel loss. During training, an output patch of specified size is randomly drawn from the training stacks at the beginning of each forward pass.

\item{\bf Class rebalancing}$\quad$
In dense output patch-based training, imbalance between the number of training samples in different classes (e.g. boundary/non-boundary) can be handled by either sampling a balanced number of pixels from an output patch, or by differentially weighting the per-pixel loss~\cite{Long2015}. In our experiment, we adopted the latter approach (loss weighting) to deal with the high imbalance between boundary and non-boundary pixels.

\item{\bf Data augmentation}$\quad$
We used the same data augmentation method used in~\cite{Ciresan2012}, randomly rotating and flipping 2D image patches.

\item{\bf Hyperparameter}$\quad$
We always used the fixed learning rate of 0.01 with the momentum of 0.9. When updating weights we divided the gradient by the total number of pixels in an output patch, similar to the typical minibatch update.
\end{enumerate}

We first trained N4 with an output patch of size $200\times200\times1$ for $90$K gradient updates. Next, we trained VD2D with $150\times150\times1$ output patches, reflecting the increased size of model compared to N4. After $60$K updates, we evaluated the trained VD2D on the training stacks to obtain preliminary boundary maps, and started training VD2D3D with $100\times100\times1$ output patches, again reflecting the increased model complexity. We trained VD2D3D for $90$K updates. In this recursive training stage we additionally used \texttt{stack4} to prevent VD2D3D from being overly dependent on the good-quality boundary maps for training stacks. It should be noted that \texttt{stack4} has slightly lower $xy$-resolution than other stacks (Table~\ref{tab:Datasets}), which we think is helpful in terms of learning multi-scale representation.

Our proposed recursive framework is different from the training of recurrent ConvNets~\cite{Pinheiro2014} in that recursive input is not dynamically produced by the first ConvNet during training, but evaluated before and being fixed throughout the recursive training stage. However, it is also possible to further train the first ConvNet even after evaluating its preliminary output as recursive input to the second ConvNet. We further trained VD2D for another $30$K updates while VD2D3D is being trained. We report the final performance of VD2D after a total of 90K updates. We also replaced the initial VD2D boundary map with the final one when evaluating VD2D3D results. With ZNNv1, it took $5$ days to train VD2D for $60$K updates, and a week to train VD2D3D for $90$K updates.

\begin{figure}[!t]
\begin{center}
\includegraphics[width=1.0\textwidth]{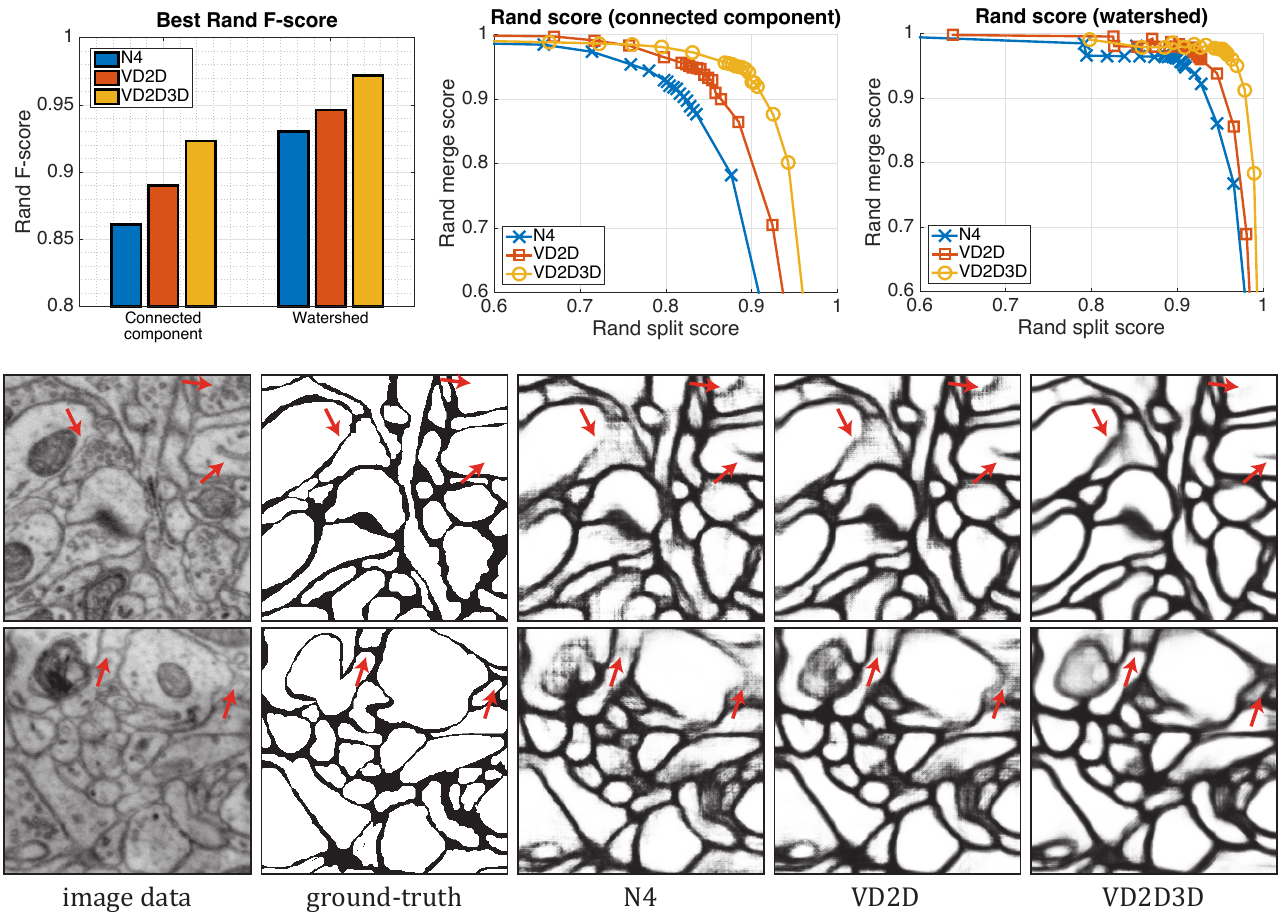}
\end{center}
\caption{Quantitative (top) and qualitative (middle and bottom) evaluation of results}
\label{results}
\end{figure}

\section{Results}
In this section, we show both quantitative and qualitative results obtained by the three architectures shown in Figure~\ref{overview}, namely N4, VD2D, and VD2D3D. The pixel-wise classification error of each model on test set was $10.63\%$ (N4), $9.77\%$ (VD2D), and $8.76\%$ (VD2D3D).

{\bf Quantitative comparison}$\quad$ Figure~\ref{results} compares the result of each architecture on test set (\texttt{stack1}), both quantitatively and qualitatively. The leftmost bar graph shows the best 2D Rand F-score of each model obtained by 2D segmentation with (1) simpler connected component clustering and (2) more sophisticated watershed-based segmentation. The middle and rightmost graphs show the precision-recall curve of each model for the Rand scores obtained with the connected component and watershed-based segmentation, respectively. We observe that VD2D performs significantly better than N4, and also VD2D3D outperforms VD2D by a significant margin in terms of both best Rand F-score and overall precision-recall curve.

{\bf Qualitative comparison}$\quad$
Figure~\ref{dataset} shows the visualization of boundary detection results of each model on test set, along with the original EM images and ground truth segmentation. We observe that false detection of boundary on intracellular regions was significantly reduced in VD2D3D, which demonstrates the effectiveness of the proposed 2D-3D ConvNet combined with recursive approach. The middle and bottom rows in Figure~\ref{results} show some example locations in test set where both 2D models (N4 and VD2D) failed to correctly detect the boundary, or mistakenly detect false boundaries, whereas VD2D3D correctly predicted on those ambiguous locations. Visual analysis on the boundary detection results of each model again demonstrates the superior performance of our proposed 2D-3D ConvNets over 2D models.

\section{Discussion}
\label{discussion}

{\bf Biologically-inspired recursive framework}$\quad$
Our proposed recursive framework is greatly inspired by the work of Chen et al.~\cite{Chen2014}. In this work, they examined the close interplay between neurons in the primary and higher visual cortical areas (V1 and V4, respectively) of monkeys performing contour detection tasks. In this task, monkeys were trained to detect a global contour pattern that consists of multiple collinearly aligned bars in a cluttered background.

The main discovery of their work is as follows: initially, V4 neurons responded to the global contour pattern. After a short time delay (${\sim}40$ ms), the activity of V1 neurons responding to each bar composing the global contour pattern was greatly enhanced, whereas those responding to the bacgkround was largely suppressed, despite the fact that those `foreground' and `background' V1 neurons have similar response properties. They referred to it as ``push-pull response mode'' of V1 neurons between foreground and background, which is attributable to the top-down influence from the higher level V4 neurons.
This process is also referred to as ``countercurrent disambiguating process''~\cite{Chen2014}.

This experimental result readily suggests a mechanistic interpretation on the recursive training of deep ConvNets for neuronal boundary detection. We can roughly think of V1 responses as lower level feature maps in a deep ConvNet, and V4 responses as higher level feature maps or output activations. Once the overall `contour' of neuonal boundaries is detected by the feedforward processing of VD2D, this preliminary boundary map can then be recursively fed to VD2D3D. This process can be thought of as corresponding to the initial detection of global contour patterns by V4 and its top-down influence on V1.

During recursive training, VD2D3D will learn how to integrate the pixel-level contextual information in the recursive input with the low-level features, presumably in such a way that feature activations on the boundary location are enhanced, whereas activations unrelated to the neuronal boundary (intracellular space, mitochondria, etc.) are suppressed. Here the recursive input can also be viewed as the modulatory `gate' through which only the signals relevant to the given task of neuronal boundary detection can pass.

Since higher level feature maps in deep ConvNets amplify signals relevant to the given task and suppress irrelevant noises through the deep composition of layers~\cite{LeCun2015}, we can expect that the effect of top-down modulation acting early on lower level feature maps will propagate through the hierarchy of layers, enhancing the signal-to-noise ratio in higher level representations to a greater extent. This is convincingly demonstrated by visualizing and comparing feature maps of VD2D and VD2D3D.

\begin{figure}[!t]
\begin{center}
\includegraphics[width=1.0\textwidth]{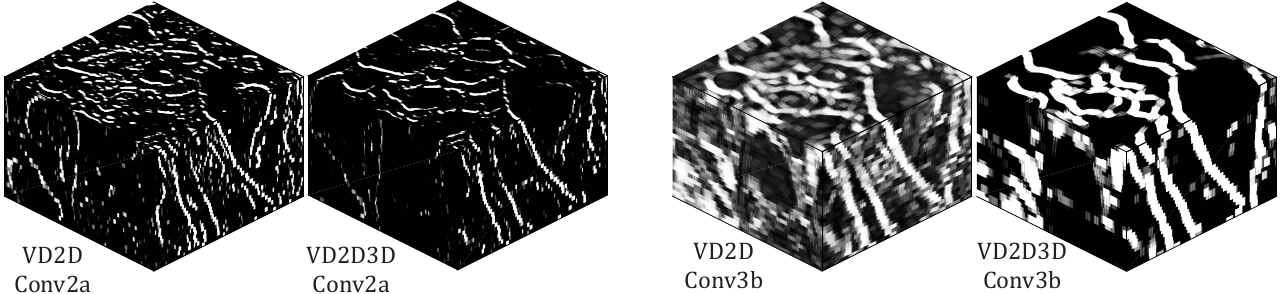}
\end{center}
\caption{Visualization of the effect of recursive training. Left: an example feature map from the layer \texttt{Conv2a} in VD2D, and its corresponding feature map in VD2D3D. Right: an example feature map from the layer \texttt{Conv3b} in VD2D, and its corresponding feature map in VD2D3D. Note that recursive training greatly enhances the signal-to-noise ratio of boundary representations.}
\label{feature}
\end{figure}

In Figure~\ref{feature},
the noisy representations of oriented boundary segments in VD2D (first and third volumes) are greatly enhanced in VD2D3D (second and fourth volumes), with signals near boundary being preserved or amplified, and noises in the background being largely suppressed. This is exactly what we expected from the proposed interpretation of our recursive framework.

{\bf Extension to 3D affinity graph}
$\quad$
Our proposed work aimed at generating the state-of-the-art 2D neuronal boundary maps that can be used as input to the 3D reconstruction pipeline, where the boundary maps are initially segmented in 2D. Contrary to this approach, Turaga et al.~\cite{Turaga2010} trained 3D ConvNets to generate 3D affinity graphs that can be directly used as input to the 3D image segmentation algorithms. Our 3D ConvNets can easily be extended to generate 3D affinity graphs by replacing the softmax output layer with the one having three independent output units, each of which generates $x$, $y$ and $z$-affinity graph, respectively.

{\bf Potential of ZNN}
$\quad$
We have shown that ZNN can serve as a viable alternative to the mainstream GPU-based deep learning frameworks, especially when processing 3D volume data with 3D ConvNets. ZNN's unique features including the large output patch-based training and the dense computation of feature maps can be further utilized for additional computations to better perform the given task. In theory, we can perform any kind of computation on the dense output prediction before each backward pass. For instance, objective functions that consider topological constraints (e.g. MALIS~\cite{Turaga2009}) or sampling of topologically relevant locations (e.g. LED weighting~\cite{Huang2014}) can be applied to the dense output patch at each gradient update.

Dense feature maps also enable the straighforward implementation of integrating multi-level features for fine-grained segmentation. Long et al.~\cite{Long2015} resorted to upsampling of the higher level features with lower resolution in order to integrate them with the lower level features with higher resolution. Since ZNN maintains every feature map at its original resolution, it is straighforward enough to combine feature maps at any level, removing the need for upsampling.

\subsubsection*{Acknowledgments}
We thank Juan C. Tapia, Gloria Choi and Dan Stettler for initial help with tissue handling and Jeff Lichtman and Richard Schalek with help in setting up tape collection. Kisuk Lee was supported by a Samsung Scholarship. The recursive approach proposed in this paper was partially motivated by Matthew J. Greene's preliminary experiments.  We are grateful for funding from the Mathers Foundation, Keating Fund for Innovation, Simons Center for the Social Brain, DARPA (HR0011-14-2-0004), and ARO (W911NF-12-1-0594).

\subsubsection*{References}

\begingroup
\renewcommand{\section}[2]{}

\endgroup



\end{document}